\documentclass[runningheads]{llncs}
\usepackage{makeidx}
\usepackage{graphicx}
\usepackage{amsmath,amssymb} % define this before the line numbering.
\usepackage{color}

\newcommand{\eqq}[1]{\begin{align}#1\end{align}}
\newcommand{\eps}{\epsilon}
\begin{document}
	\pagestyle{headings}
	\mainmatter

	% Insert your submission number here
	\def\GCPR17SubNumber{60}

	% Replace with your title
%	\title{Semantic Boundary Detection Without Pixel-wise Labels}
	\title{Object Boundary Detection and Classification with Image-level Labels}

	% DO NOT MODIFY these for the draft version that is used for the
	% review process.
	\titlerunning{Object Boundary Detection and Classification with Image-level Labels}
	\authorrunning{JY Koh, W Samek, KR M{\"u}ller, A Binder}
	\author{Jing Yu Koh$^1$, Wojciech Samek$^2$, Klaus-Robert M{\"u}ller$^{3,4}$, Alexander Binder$^1$}
	\institute{$^{1}$ISTD Pillar, Singapore University of Technology and Design, Singapore\\$^{2}$Department of Video Coding \& Analytics, Fraunhofer Heinrich Hertz Institute, Berlin, Germany\\$^{3}$Department of Computer Science, TU Berlin, Germany\\$^4$Department of Brain and Cognitive Engineering, Korea University, Seoul, Republic of Korea}

	\maketitle

%===========================================================
\begin{abstract}
Semantic boundary and edge detection aims at simultaneously detecting object edge pixels in images and assigning class labels to them. Systematic training of predictors for this task requires the labeling of edges in images which is a particularly tedious task. We propose a novel strategy for solving this task, when pixel-level annotations are not available, performing it in an almost zero-shot manner by relying on conventional whole image neural net classifiers that were trained using large bounding boxes. Our method performs the following two steps at test time. Firstly it predicts the class labels by applying the trained whole image network to the test images. Secondly, it computes pixel-wise scores from the obtained predictions by applying backprop gradients as well as recent visualization algorithms such as deconvolution and layer-wise relevance propagation. We show that high pixel-wise scores are indicative for the location of semantic boundaries, which suggests that the semantic boundary problem can be approached without using edge labels during the training phase.
\end{abstract}

%===========================================================
\section{Introduction}

Neural net based predictors achieve excellent results in many data-driven tasks, examples among the newer being \cite{DBLP:conf/cvpr/HariharanAGM15,DBLP:conf/iccv/MalinowskiRF15,DBLP:conf/cvpr/LongSD15,DBLP:conf/gecco/KoutnikCSG13,mnih-dqn-2015}, while others such as video detection or machine translation \cite{DBLP:conf/nips/SutskeverVL14,DBLP:conf/cvpr/KarpathyTSLSF14} are equally impressive. Rather than extending neural networks to a new application, we focus here on the question whether a neural network can solve problems which are {\it harder} than the one for which the network was trained. In particular, we consider the task of semantic boundary detection which we aim to solve without appropriately fine-grained training labels. 
 %the most famous being image classification \cite{DBLP:conf/nips/KrizhevskySH12}, image detection \cite{DBLP:journals/corr/GirshickDDM13} and machine translation \cite{DBLP:conf/nips/SutskeverVL14}.

The problem of semantic boundary detection (SBD) \cite{DBLP:conf/iccv/HariharanABMM11} can be defined as the simultaneous detection of object edge pixels and assignment of class labels to such edge pixels in images. Recently, the work of \cite{DBLP:conf/iccv/BertasiusST15,7917294,Yang_2016_CVPR,2017arXiv170509759Y} showed substantial improvement using neural nets, however, the approach relied on end-to-end training with a dataset for which semantic boundary labels were available. 
When trying to build a predictor for SBD, practitioners face the problem that the classical inductive machine learning paradigm requires to create a dataset with semantic boundary labels, that is, for each image a subset of pixels in images corresponding to object edges is labeled with class indices. Creating such labelling is a particularly tedious task, unlike labelling whole images or drawing bounding boxes, both of which can be done very quickly. The best proof for this difficulty is the fact that we are aware of only one truly semantic boundary dataset \cite{DBLP:conf/iccv/HariharanABMM11}. %Segmentation is able to yield approximate ground truth when followed by morphological operators, but this faces problems with occlusions and clutter which create additional boundaries.
%Another task related to SBD is the labelling of object edges with class labels in images. We will refer to this task as semantic edge detection (SED). The difference between SBD and SED is that in SED object edges may appear inside an object. Note that SBD and SED are different from contour detection tasks which intend to find contours of objects without assigning class labels to them. 
Note that SBD is different from contour detection tasks \cite{Xie2017} which aim at finding contours of objects without assigning class labels to them. In that sense the scope of our proposed work is different from unsupervised contour detection as in \cite{Li_2016_CVPR}.

The main question in this paper is to what extent it is possible to solve the semantic boundary or edge detection task without having appropriately fine-grained labels, i.e., pixel-level ground truth, which are required for classical training paradigms ?  We do not intend to replace the usage of pixel-wise boundary labels when they are available. We aim at use cases in which pixel-wise boundary labels are not available during the training phase. One example of using weaker annotations for semantic boundary detection is \cite{Khoreva_2016_CVPR} where bounding box labels are used to learn semantic boundaries. We propose a novel strategy to tackle a problem requiring fine-grained labels, namely semantic boundary detection, with a classifier trained for image classification using only image-wise labels. 
For that we use neural nets that classify an image, and apply existing visualization methods that are able to assign class-specific-scores to single pixels. These class-specific pixel scores can then be used to define semantic boundary predictions.

The contribution of this paper is as follows. We demonstrate that classifier visualization methods are useful beyond producing nice-to-look-at images, namely for approaching prediction tasks on the pixel-level in the absence of appropriately fine-grained training labels. As an example, we apply and evaluate the performance of classifier visualization methods to the SBD task. We show that these visualization methods can be used for producing quantifiably meaningful predictions at a higher spatial resolution than the labels, which were the basis for training the classifiers. We discuss the shortcomings of such approaches when compared to the proper training paradigm that makes use of pixel-level labels. We do not expect such methods to beat baselines that employ the proper training paradigm and thus use pixel-level labels during training, but rather aim at the practitioner's case in which fine-grained training data is too costly in terms of money or time.

%%%%%%%%%%%%%%%%%%%%%%%%%
\section{Obtaining Pixel-level Scores from Image-wise Predictions}
In the following we introduce the methods that we will use for producing pixel-level scores without pixel-level labels during training time. It is common to all these methods that they take a classifier prediction $f_c(x)$ on an image $x$ and produce scores $s_c(p)$ for pixels $p \in x$. Suppose we have classifiers $f_c(x)$ for multiple classes $c$. 
Then we can tackle the SBD problem by (1) classifying an image, i.e., determine those classes that are present in the image, and (2) computing pixel-wise scores for those classes using one of the following methods.

\subsection{Gradient}
Probably the most obvious idea to tackle the SBD problem is to run a forward prediction with a classifier, and compute the gradient for each pixel. Let $x$ be an input image, $f_1,\ldots,f_C$ be $C$ outputs of a multi-class classifier and $x_p$ be the $p$-th pixel. Computing pixel-wise scores for a class $c$ and pixel $p$ can be achieved using
\eqq{
s(p) & = \left\| \frac{\partial f_c}{\partial x_p}(x)\right\|_2
}
The norm runs here over the partial derivatives for the $(r,g,b)$-subpixels of a pixel $p$. Alternatively one can sum up the subpixel scores in order to have a pixel-score.
Using gradients for visualizing sensitivities of neural networks has been shown in \cite{DBLP:journals/corr/SimonyanVZ13b}. A high score in this sense indicates that the output $f_c$ has high sensitivity under small changes of the input $x_p$, i.e. there exists a direction in the tangent space located at $x$ for which the slope of the classifier $f_c$ is very high.

In order to see the impact of partial derivatives, consider the case of a simple linear mapping that takes subpixels $x_{p,s}$ of pixel $p$ as input.
\eqq{
f(x) & = \sum_{p } \sum_{s \in \{r,g,b\}} w_{p,s} x_{p,s} 
}
In this case backpropagation combined with an $\ell_2$-norm yields:
\eqq{
s(p)= ( w_{p,r}^2+w_{p,g}^2+w_{p,b}^2  )^{1/2}
}
Note that the input $x_{p,s}$, and in particular its sign plays no role in a visualization achieved by backpropagation, although obviously the sign of $x_{p,s}$ does matter for deciding whether to detect an object ($f(x)>0$) or not ($f(x)<0$). This is a limiting factor, when one aims to explain what pixels are relevant for the prediction $f(x)>0$.

\subsection{Deconvolution}
Deconvolution \cite{DBLP:conf/eccv/ZeilerF14} is an alternative method to compute pixel-wise scores. Starting with scores given at the top of a convolutional layer, it applies the transposed filter weights to compute scores at the bottom of the same layer. Another important feature is used in max-pooling layers, where scores from the top are distributed down to the input that yielded the maximum value in the max pooling. 
Consider the linear mapping case again. Then deconvolution in the sense of multiplying the transposed weights $w$ (as it is for example implemented in the Caffe package) yields for subpixel $s$ of channel $p$ 
\eqq{
s(p,s)&= f_c(x)w_{p,s}
}
This score can be summed across subpixels, or one can take again an $\ell_p$-norm. When using summation across subpixels, then deconvolution is proportional to the prediction $f_c(x)$, in particular it expresses the dominating terms $w_{p,s} x_{p,s} \approx f_c(x)$ correctly which contribute most to the prediction $f(x)$.

\subsection{Layer-wise Relevance Propagation}
Layer-wise Relevance Propagation (LRP) \cite{BacBinMonKlaMueSam15} is a principled method for explaining neural network predictions in terms of pixel-wise scores. LRP reversely propagates a numerical quantity, named relevance, in a way that preserves the sum of the total relevance at each layer.
The relevance is initialized at the output as the prediction score $f_c(x)$ and propagated down to the inputs (i.e., pixels), so that the relevance conservation property holds at each layer
\begin{eqnarray}
f_c(x) = \ldots = \sum_j R_j^{(l+1)} \ldots = \sum_i R_i^{(l)} = \ldots  = \sum_p R_p^{(1)}
\end{eqnarray}
where $\{R_j^{(l+1)}\}$ and $\{R_i^{(l)}\}$ denote the relevance at layer $l+1$ and $l$, respectively, and $\{R_p^{(1)}\}$ represents the pixel-wise relevance scores.

Let us consider the neural network as an feed-forward graph of elementary computational units (neurons), each of them realizing a simple function of type
\begin{eqnarray}
x_j^{(l+1)} = g\Big(0,\sum_i x_i^{(l)} w_{ij}^{(l,l+1)} + b_j^{(l+1)}\Big)\quad 
\text{ e.g. }\ g(z)=\max(0,z)
\end{eqnarray}
where $j$ denotes a neuron at a particular layer $l+1$, and, where $\sum_i$ runs over all lower-layer neurons connected to neuron $j$. $w_{ij}^{(l,l+1)}, b_j^{(l+1)}$ are parameters of a neuron. The prediction of a deep neural network is obtained by computing these neurons in a feed-forward pass. Conversely, \cite{BacBinMonKlaMueSam15} have shown that the same graph structure can be used to redistribute the relevance $f(x)$ at the output of the network onto pixel-wise relevance scores $\{R_p^{(1)}\}$, by using a local redistribution rule
\begin{eqnarray}
R_i^{(l)} = \sum_j \frac{z_{ij}}{\sum_{i'} z_{i'j}} R_j^{(l+1)} \quad \rm{with}\quad z_{ij} = x_i^{(l)} w_{ij}^{(l,l+1)}
\label{eq:LRPnaive}
\end{eqnarray}
where $i$ indexes a neuron at a particular layer $l$, and where $\sum_j$ runs over all upper-layer neurons to which neuron $i$ contributes. 
Application of this rule in a backward pass produces a relevance map (heatmap) that satisfies the desired conservation property. % $\sum_p R_p^{(1)} = f(\boldsymbol{x})$. 
%This decomposition algorithm is termed Layer-wise Relevance Propagation (LRP). %See Fig.~\ref{fig:overview} for an overview.

We consider two other LRP algorithms introduced in \cite{BacBinMonKlaMueSam15}, namely the $\epsilon$-variant and the $\beta$-variant. The first rule is given by:
\begin{align}
R_{i}^{(l)} = \sum_{j} \frac{z_{ij}}{\sum_{i'} z_{i'j}+\epsilon \,\mathrm{sign}(\sum_{i'} z_{i'j}) } R_j^{(l+1)} \label{eq:lrp-basic}
\end{align}
Here for $\epsilon > 0$ the conservation idea is relaxated in order to gain better numerical stability. The second formula is given by:
\begin{align}
R_{i}^{(l)} = \sum_{j} \left( \alpha \cdot \frac{z_{ij}^+}{\sum_{i'} z_{i'j}^+} + \beta \cdot \frac{z_{ij}^-}{\sum_{i'}z_{i'j}^-} \right) R^{(l+1)}_j \label{eq:lrp-alphabeta}.
\end{align}
Here, $z_{ij}^+$ and $z_{ij}^-$ denote the positive and negative part of $z_{ij}$ respectively, such that $z_{ij}^+ + z_{ij}^- = z_{ij}$. We enforce $\alpha + \beta = 1$, $\alpha >0$, $\beta \le 0$ in order for the relevance propagation equations to be conservative layer-wise. Note that for $\alpha = 1$ this redistribution rule is equivalent (for ReLU nonlinearities $g$) to the $z^+$-rule by \cite{MonArXiv15}.

In contrast to the gradient, LRP recovers the natural decomposition of a linear mapping \eqq{f(x) = \sum_{i=1}^D w_i x_i} i.e., the pixel-level score \eqq{R_i = w_i x_i \label{eq:lin}} not only depends on whether the classifier reacts to this input dimension ($w_i > 0$), but also if that feature is actually present ($x_i > 0$).
An implementation of LRP can be found in \cite{LapJMLR16}.

%%%%%%%%%%%%%%%%%%%%%%%%%
\section{Experiments}
We perform the experiments on the SBD dataset with a Pascal VOC multilabel classifier from \cite{BacCVPR16} that is available in the BVLC model zoo of the Caffe \cite{DBLP:conf/mm/JiaSDKLGGD14} package. This classifier was trained using the 4 edge crops and the center crops of the ground truth bounding boxes of the Pascal VOC dataset \cite{voc09report}. We do not use pixel labels at training time, however, for evaluation at test time we use the pixel-wise ground truth, in order to be able to compare all methods quantitatively. Same as \cite{DBLP:conf/iccv/HariharanABMM11} we report the maximal F-score and the average precision on the pixel-level of an image. We stick to the same convention regarding counting true positives in a neighborhood, as introduced in \cite{DBLP:conf/iccv/HariharanABMM11}. 

\subsection{Performance on the SBD task}
Table \ref{tab:ap-sbd} shows the average precision (AP) scores for all methods.
\begin{table}
\centering
\caption{\label{tab:ap-sbd} Average precision (AP) and maximal F-scores (MF) scores of various methods to compute pixel-wise scores from whole image classifiers without pixel-labels at training time, compared against the original method \emph{InverseDetectors} \cite{DBLP:conf/iccv/HariharanABMM11} and Boundary detection using Neural nets \emph{HFL} \cite{DBLP:conf/iccv/BertasiusST15}. Only the last two both use pixel-labels at training time.  All other use no pixel-level labels during training. Grad denotes Gradient, Deconv denotes \cite{DBLP:conf/eccv/ZeilerF14}, $\eps$ and $\beta$ refer to LRP variants given in equations \eqref{eq:lrp-basic} and \eqref{eq:lrp-alphabeta} taken from \cite{BacBinMonKlaMueSam15}.}
\begin{tabular}{ccccccc|cc}\hline
training phase: & \multicolumn{6}{c|}{image-level labels} & \multicolumn{2}{c}{pixel-level labels} \\
Method: & Gradient & Deconv & $\beta=0$ & $\beta=-1$ & $\eps=1$ & $\eps=0.01$ & InvDet \cite{DBLP:conf/iccv/HariharanABMM11} & HFL \cite{DBLP:conf/iccv/BertasiusST15} \\ \hline
AP & 22.5 & 25.0 & 28.4    & 27.3 & \textbf{31.4} & 31.2 & 19.9 & \textbf{54.6} \\
MF & 31.0 & 33.3  & 35.1  & 34.1  &     38.0  & \textbf{38.1} & 28.0 & \textbf{62.5} \\\hline
\end{tabular}
\vspace{1mm}
\end{table}
We can see from the table that the neural-network based method \cite{DBLP:conf/iccv/BertasiusST15} which uses pixel-level ground truth at training time performs best by a large margin. Methods that do not employ pixel-level labels at training time perform far worse. However, we can see a certain surprise: all the methods perform better than the method \cite{DBLP:conf/iccv/HariharanABMM11} on Semantic Boundary Detection that was the best baseline before the work of \cite{DBLP:conf/iccv/BertasiusST15} replaced it. 
Note that \cite{DBLP:conf/iccv/HariharanABMM11} as well as \cite{DBLP:conf/iccv/BertasiusST15} relies on pixel-wise labels during training, whereas the proposed methods require only image-wise labels.
This result gives a realistic comparison of how good methods on pixel-wise prediction without pixel-labels in the training phase can perform.

The pixel-wise scores for LRP are computed by summing over subpixels. For Gradient and Deconvolution using the negative sum over subpixels performed better than using the sum or the $\ell_2$-norm. For both cases negative pixel scores were set to zero. This follows our experience with Deconvolution and LRP that wave-like low-level image filters, which are typically present in deep neural nets, receive equally wave-like scores with positive and negative half-waves. Removing the negative half waves improves the prediction quality.
\begin{table}
\centering
\caption{\label{tab:cmp-subpixels} Comparison of various ways to combine subpixel scores into a pixel-wise score.}
\begin{tabular}{cp{0.5cm}cp{0.5cm}cp{0.5cm}c}\hline
subpixel aggregation method & & sum  & & sum of negative scores & & $\ell_2$-norm \\
Gradient AP & & 22.0 & & 22.5 & & 18.8 \\
Deconvolution AP & & 22.9 & & 25.0 & & 21.9 \\ \hline
\end{tabular}
\vspace{1mm}
\end{table}
Table \ref{tab:cmp-subpixels} shows the comparison of AP scores for various methods to compute a pixel-wise score from subpixel scores. Note that we do not show the $\ell_2$-norm, or the summed negative scores for the LRP methods, as LRP does preserve the sign of the prediction and thus using the sum of negative scores or $\ell_2$-norm has no meaningful interpretation for LRP.

\subsection{Shortcomings of visualization methods}

\begin{figure}
\begin{center}
\begin{tabular}{p{0.16\textwidth}p{0.16\textwidth}p{0.16\textwidth}p{0.16\textwidth}p{0.16\textwidth}p{0.16\textwidth}}
image & ground truth  & gradient & Deconv & LRP-$\beta=0$ & LRP-$\eps=1$
\end{tabular}
\includegraphics[width=0.16\textwidth]{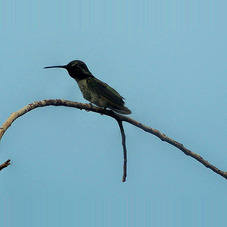}
\includegraphics[width=0.16\textwidth]{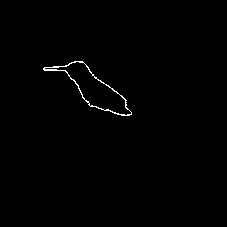}
\includegraphics[width=0.16\textwidth]{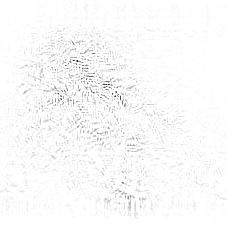}
\includegraphics[width=0.16\textwidth]{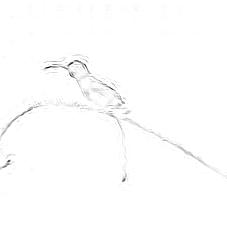}
\includegraphics[width=0.16\textwidth]{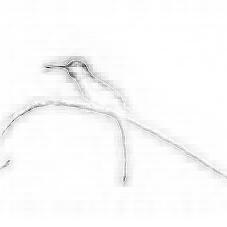}
\includegraphics[width=0.16\textwidth]{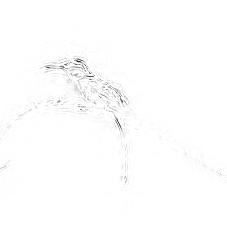}\\
\includegraphics[width=0.16\textwidth]{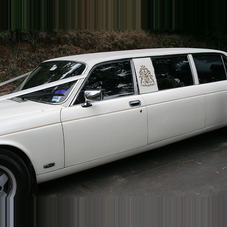}
\includegraphics[width=0.16\textwidth]{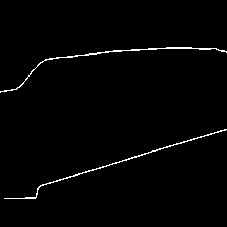}
\includegraphics[width=0.16\textwidth]{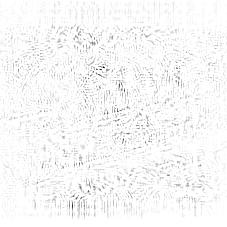}
\includegraphics[width=0.16\textwidth]{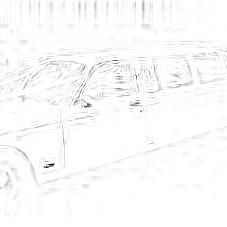}
\includegraphics[width=0.16\textwidth]{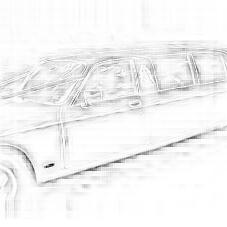}
\includegraphics[width=0.16\textwidth]{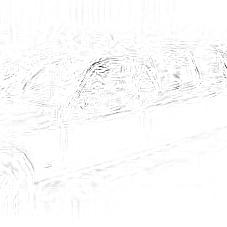}\\
\includegraphics[width=0.16\textwidth]{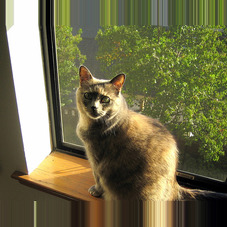}
\includegraphics[width=0.16\textwidth]{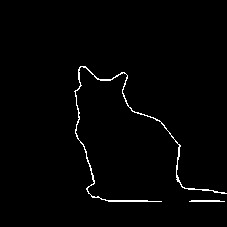}
\includegraphics[width=0.16\textwidth]{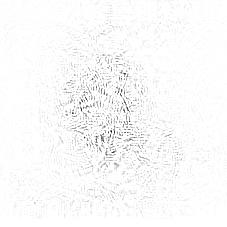}
\includegraphics[width=0.16\textwidth]{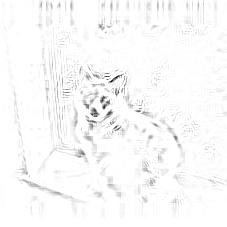}
\includegraphics[width=0.16\textwidth]{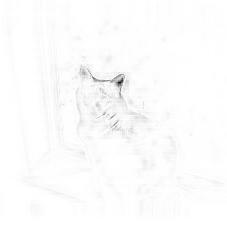}
\includegraphics[width=0.16\textwidth]{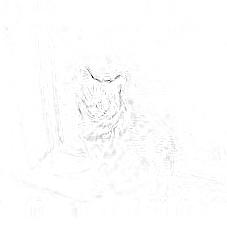}\\
\includegraphics[width=0.16\textwidth]{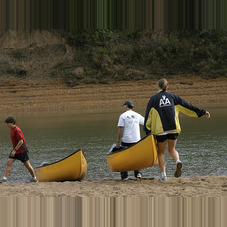}
\includegraphics[width=0.16\textwidth]{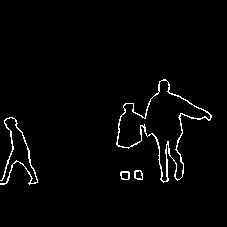}
\includegraphics[width=0.16\textwidth]{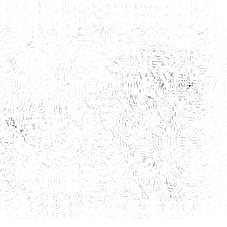}
\includegraphics[width=0.16\textwidth]{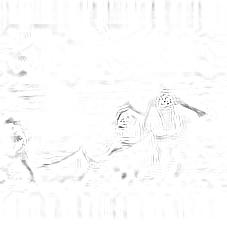}
\includegraphics[width=0.16\textwidth]{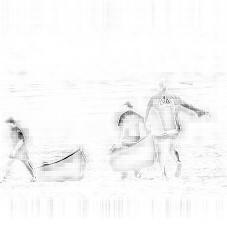}
\includegraphics[width=0.16\textwidth]{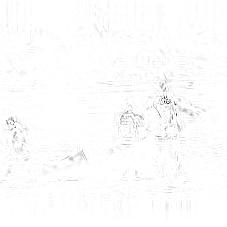}\\
\includegraphics[width=0.16\textwidth]{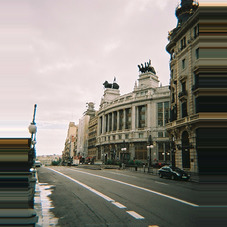}
\includegraphics[width=0.16\textwidth]{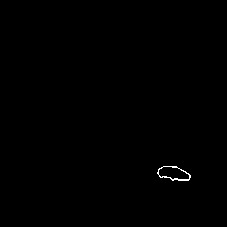}
\includegraphics[width=0.16\textwidth]{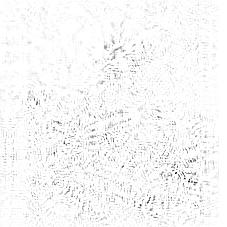}
\includegraphics[width=0.16\textwidth]{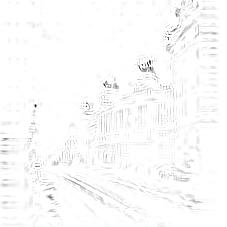}
\includegraphics[width=0.16\textwidth]{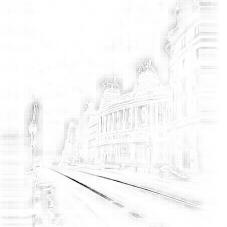}
\includegraphics[width=0.16\textwidth]{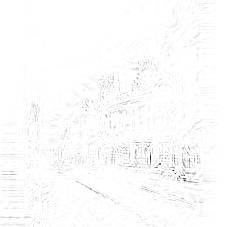}\\
\includegraphics[width=0.16\textwidth]{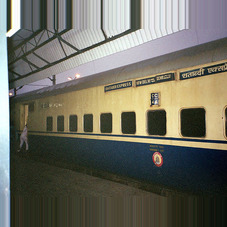}
\includegraphics[width=0.16\textwidth]{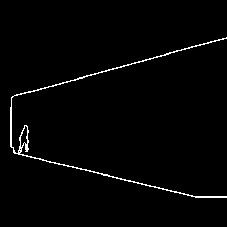}
\includegraphics[width=0.16\textwidth]{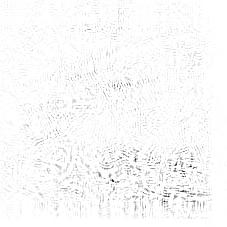}
\includegraphics[width=0.16\textwidth]{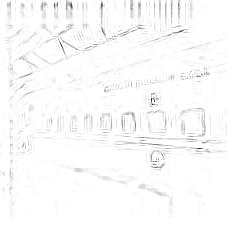}
\includegraphics[width=0.16\textwidth]{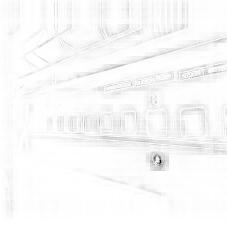}
\includegraphics[width=0.16\textwidth]{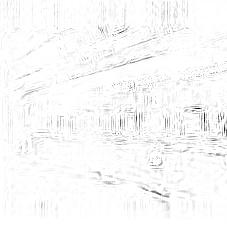}\\
\end{center}
\caption{\label{fig:gtvshm}Heatmaps of pixel-wise scores compared against the groundtruth. From left to right: original image, pixel-level ground truth, gradient (negative scores), Deconvolution (negative scores), LRP with $\beta=0$ and with $\eps=1$.}
\end{figure}

\begin{figure}
\begin{center}
\includegraphics[width=0.18\textwidth]{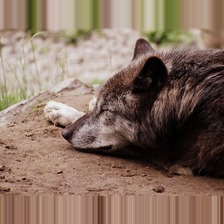}\includegraphics[width=0.18\textwidth]{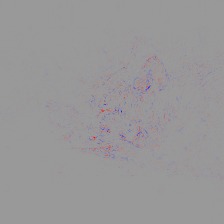}\includegraphics[width=0.18\textwidth]{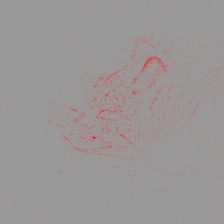}\\
\includegraphics[width=0.18\textwidth]{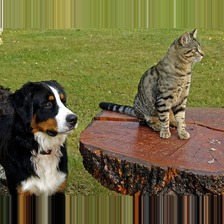}\includegraphics[width=0.18\textwidth]{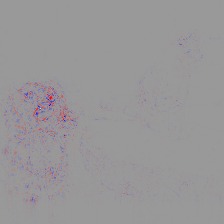}\includegraphics[width=0.18\textwidth]{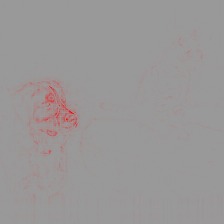}\\
\includegraphics[width=0.18\textwidth]{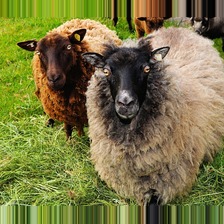}\includegraphics[width=0.18\textwidth]{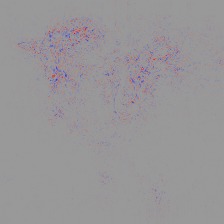}\includegraphics[width=0.18\textwidth]{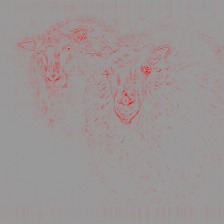}\\
\end{center}
\caption{\label{label:figggm}Heatmaps of pixel-wise scores computed for the GoogleNet Reference Classifier of the Caffe Package show the sparsity of pixel-wise prediction methods. The used classifiers were: Timber wolf, Bernese mountain Dog and Ram. Left Column: image as it enters the deep neural net. Middle: pixel-wise scores computed by LRP with $\eps=1$. Right: pixel-wise scores computed by LRP with $\beta=0$.}
\end{figure}

Semantic boundaries are not the most relevant regions for the decision of above mentioned classifiers trained on images of natural scenes. This does not devaluate models trained on shapes. It merely says that, given RGB images of natural scenes as input, above object class predictors put considerable weight on internal edges and textures rather than outer boundaries, an effect which can be observed in the heatmaps in Figures \ref{fig:gtvshm} and \ref{label:figggm}. This is the primary hypothesis why the visualization methods above are partially mismatching the semantic boundaries. We demonstrate this hypothesis quantitatively by an experiment. For this we need to introduce a measure of relevance of a set of pixels which is independent of the computed visualizations.

\paragraph{Perturbation analysis}
We can measure the relevance of a set of pixels $\mathcal{S} \subset x$ of an image $x$ for the prediction of a classifier by replacing the pixels in this set by some values, and comparing the prediction on the modified image $\tilde{x}_{\mathcal{S}}$ against the prediction score $f(x)$ for the original image \cite{samek2015evaluating} (similar approach has been applied for text in \cite{ArrACL16}). This idea follows the intuition that most random perturbations in a region that is important for classification will lead to a decline of the prediction score for the image as a whole: $f(\tilde{x}_{\mathcal{S}}) < f(x)$. It is clear that there exist perturbations of a region yield an increase of the prediction score: for example a change that follows the gradient direction locally. Thus we will draw many perturbations of the set $\mathcal{S}$ from a random distribution $P$ and measure an approximation the expected decline of the prediction
\eqq{
m= f(x)-\mathbb{E}_{\mathcal{S} \sim P } [f(\tilde{x}_{\mathcal{S}} )] \label{eq:decavg} 
}
We intend to measure the expected decrease for the set $\mathcal{S}$ being the ground truth pixels for the SBD task, and compare it against the set of highest scoring pixels. For a fair comparison the set of highest scoring pixels will be limited to have the same size as the number of ground truth pixels. Highest scoring pixels will be defined by the pixel-wise scores from the above methods. We will show that the expected decrease is higher for the pixel-wise scores, which indicates that ground truth pixels representing semantic boundaries are not the most relevant subset for the classifier prediction.

The experiment to demonstrate this will be designed as follows. For each test image and each ground truth class we will take the set of ground truth pixels, and randomly perturb them. For a $(r,g,b)$-pixel we will draw the values from a uniform distribution in $[0,1]^3 \subset \mathbb{R}^3$.
For each image and present class of semantic boundary task ground truth we repeat $200$ random perturbations of the set in order to compute an approximation to Equation \eqref{eq:decavg}. We compute the average over all images to obtain the average decrease on ground truth pixels $m_{GT}$. $m_{GT}$ is an average measure of relevance of the ground truth pixels.
$m_{GT}$ is to be compared against the analogous quantity $m_V$ derived from the top-scoring pixels of a visualization method. For a given visualization method $V \in \{Gradient, Deconv, LRP$-$\beta$, $LRP$-$\eps\}$, we define the set of pixels to be perturbed as the pixels with the highest pixel-wise scores computed from the visualization method. The set size for this set will be the same as the number of ground truth pixels of the semantic boundary task of the same image and class. Running the same perturbation idea according to Equation \eqref{eq:decavg} on this set yields a measure $m_{V}$ of average decrease of classifier prediction that is specific to the most relevant pixels of the given visualization method.

\begin{table}
\centering
\caption{\label{tab:cmp-perturb} Comparison of the averaged prediction scores. $f(x)$ denotes the average prediction for the unperturbed images for all ground truth classes. $m_{GT}$ denotes the average prediction for images with perturbed ground truth pixels. $m_{Deconv}$ and $m_{LRP,\eps=1}$ denotes the average prediction for images with perturbed highest scoring pixels having the same cardinality as the ground truth pixels, using Deconvolution and LRP.}
\begin{tabular}{cp{0.25cm}cp{0.25cm}cp{0.25cm}cp{0.25cm}c}\hline
 $f_c(x)$ & & $m_{GT}$ & & $m_{Deconv}$ & & $m_{LRP,\eps=1}$ \\\hline
 10.20 $\pm$ 0 & & 7.73 $\pm$ 0.36 & & 5.68 $\pm$ 0.38 & & 1.73 $\pm$ 0.34  \\\hline
\end{tabular}
\vspace{1mm}
\end{table}

Table \ref{tab:cmp-perturb} shows the results of the comparison. Note that we take the ground truth in the image that has been resized to match the receptive field of the neural net ($227 \times 227$), and apply one step of classical morphological thickening. This thickened ground truth set will be used.  The standard deviation was computed for the $200$ random perturbations and averaged over images and classes. We can see from the table that the decrease is stronger for the visualization methods compared to the ground truth pixels. This holds for Deconvolution as well as for LRP. The pixels highlighted by these methods are more relevant for the classifier prediction, even though they disagree with boundary pixel labels. In summary this supports our initially stated hypothesis that boundary pixels are not the most relevant for classification, and our explanation why these methods are partially mismatching the set of boundary ground truth labels.

We can support this numerical observation also by example images. We can observe two error cases. Firstly, the pixel-wise predictions may miss semantic boundaries that are deemed to be less discriminative for the classification task. This adds to false negatives. Secondly, the pixel-wise predictions may assign high scores to pixels that are relevant for the classification of an object and lie inside the object. 
Figure \ref{fig:gtvshm} shows some examples. We can clearly see false negatives and false positives in these examples, for example for the car and LRP-$\eps=1$ where the window regions are deemed to be highly relevant for the classifier decision, but the outer boundary on the car top is considered irrelevant which is a bad result with respect to boundary detection. For the cat most of the methods focus on its face rather than the cat boundaries. The bird is an example where deconvolution gives a good result. For the people with the boats the heatmap is shown for the people class. In this example LRP-$\eps=1$ focuses correctly most selectively on the people, same as for the tiny car example.

We can observe from these figures a common sparsity of the pixel-wise prediction methods. This motivates why we did not aim at solving segmentation tasks with these methods.
Finally we remark that this sparsity is not an artefact of the particular deep neural network from \cite{BacCVPR16} tuned for PASCAL VOC. Figure \ref{label:figggm} shows the same effect for the GoogleNet Reference Classifier \cite{DBLP:journals/corr/SzegedyLJSRAEVR14} of the Caffe Package \cite{DBLP:conf/mm/JiaSDKLGGD14}. As an example, for the wolf, parts of the body in the right have missing boundaries. Indeed this part is not very discriminative. A similar interpretation can be made for the lower right side of the dog which has a strong image gradient but not much dog-specific evidence. 

%%%%%%%%%%%%%%%%%%%%%%%%%
\section{Conclusion}
We presented here several methods for zero-shot learning for semantic boundary detection and evaluated them quantitatively. These methods are useful when pixel-level labels are unavailable at training time. These methods perform reasonably against previous state of the art. It would be interesting to evaluate these methods on other datasets with class-specifically labeled edges, if they would become available in the future. Furthermore we have shown that classifier visualization methods \cite{DBLP:journals/corr/SimonyanVZ13b,DBLP:conf/eccv/ZeilerF14,BacBinMonKlaMueSam15} have applications beside pure visualization due to their property of computing predictions at a finer scale.

\bibliographystyle{splncs03}
\bibliography{references}

%this would normally be the end of your paper, but you may also have an appendix
%within the given limit of number of pages
\end{document}